\pdfoutput=1

\documentclass[11pt]{article}

\usepackage{authblk}
\usepackage{EMNLP2022}

\usepackage{times}
\usepackage{latexsym}
\usepackage{booktabs}
\usepackage{multirow}
\usepackage{graphicx}
\usepackage{mathtools}
\usepackage{cleveref}
\usepackage{comment}
\usepackage{subfigure}

\usepackage[T1]{fontenc}

\usepackage[utf8]{inputenc}

\usepackage{microtype}

\usepackage{inconsolata}

\newlength{\bibitemsep}
\newlength{\bibparskip}
\let\oldthebibliography\thebibliography
\renewcommand\thebibliography[1]{%
  \oldthebibliography{#1}%
  \setlength{\parskip}{\bibitemsep}%
  \setlength{\itemsep}{\bibparskip}%
}

\title{On Task-Adaptive Pretraining for Dialogue Response Selection}

\author{\textbf{Tzu-Hsiang Lin}}
\author[1]{\textbf{Ta-Chung Chi}}
\author[2]{\textbf{Anna Rumshisky}}
\affil[1]{Language Technologies Institute, Carnegie Mellon University}
\affil[2]{Department of Computer Science, University of Massachusetts Lowell}
\affil[ ]{\texttt{tzuhsial@alumni.cmu.edu,tachungc@cmu.edu,arum@cs.uml.edu}}

\begin{document}
\maketitle
\begin{abstract}

Recent advancements in dialogue response selection (DRS) 
are based on the \textit{task-adaptive pre-training (TAP)} approach, by first initializing their model with BERT~\cite{devlin-etal-2019-bert}, and adapt to dialogue data with dialogue-specific or fine-grained pre-training tasks.  
However, it is uncertain whether BERT is the best initialization choice, or whether the proposed dialogue-specific fine-grained learning tasks are actually better than MLM+NSP.
This paper aims to verify assumptions made in previous works and understand the source of improvements for DRS.
We show that initializing with RoBERTa achieve similar performance as BERT, and MLM+NSP can outperform all previously proposed TAP tasks, during which we also contribute a new state-of-the-art on the Ubuntu corpus.
Additional analyses shows that the main source of improvements comes from the TAP step, and that the NSP task is crucial to DRS, different from common NLU tasks.

\end{abstract}

\section{Introduction}

\begin{table*}[h]
    \centering
    \scriptsize 
    \begin{tabular}{c|c c c c|c c c c|c c c c}
    \toprule 
       \multirow{2}{*}{Dataset}  & \multicolumn{4}{c}{Ubuntu} & \multicolumn{4}{c}{Douban} & \multicolumn{4}{c}{E-commerce} \\
                & Pre-train  & Train & Valid & Test & Pre-train  & Train & Valid & Test & Pre-train  & Train & Valid & Test  \\
      \midrule 
       \# pairs & 5.1m & 1m & 500k & 500k & 3.3m & 1m & 50k & 50k & 2.8m & 1m & 50k & 50k\\
       pos:neg  & 1:1 & 1:1 & 1:9 & 1:9 & 1:1 & 1:1 & 1:1 & 1.2:8.8 & 1:1 & 1:1 & 1:1 & 1:9 \\
       \# avg turns & 7.48 & 10.13 & 10.11 & 10.11 & 5.36 & 6.69 & 6.75 & 6.48 & 6.31 & 5.51 & 5.48 & 5.64  \\
       \bottomrule
    \end{tabular}
    \caption{Dataset statistics.  Pre-train set is generated from train set using the method in ~\Cref{subsubsec:data_aug}.}
    \label{tab:datasets}
    \vspace{-15pt}
\end{table*}

Recent advances in dialogue response selection (DRS)~\cite{wang-etal-2013-dataset,alrfou2016conversational} have mostly adopted the \textit{Task-adaptive Pre-training (TAP)} approach~\cite{gururangan-etal-2020-dont} and can be divided into three steps.  Step one: initialize the model using a pre-trained language model checkpoint.  Step two: perform TAP on the DRS training data with data augmentation.  Step three: fine-tune the TAP on the DRS dataset.  

For step one, surprisingly, all previous works have exclusively chosen the original BERT~\cite{devlin-etal-2019-bert}.  We hypothesize this is due to the \textit{Next Sentence Prediction (NSP)} task in BERT having the same formulation as DRS, and previous works have thus assumed BERT contains more knowledge related to DRS.
Nonetheless, it is well-known in NLP literature that BERT is under-trained and that removing the NSP task during pre-training improves downstream task performance~\cite{roberta}.
For step two, while earlier work uses MLM+NSP for TAP~\cite{whang20_interspeech, SA-BERT}, more recent works have assumed that MLM+NSP is too simple or does not directly model dialogue patterns, and thus proposed various dialogue-specific learning tasks~\cite{whang2021ums,Xu2021LearningAE} or fine-grained pre-training objectives to help the DRS models better learn dialogue patterns and granular representations~\cite{su-etal-2021-dialogue}.  
However, ~\citet{han-etal-2021-fine} recently uses MLM with a simple variant of NSP for TAP and outperforms almost all of them, raising questions on whether these dialogue-specific fine-grained learning tasks are actually better.

This paper aims to verify the assumptions made in previous works and understand the source of improvements from the TAP approach. 
First, we include RoBERTa~\cite{roberta} as an additional initialization checkpoint to BERT and use MLM+NSP for TAP.
Experiments on Ubuntu, Douban and E-commerce benchmarks show that (1) BERT and RoBERTa performs similarly, and (2) MLM+NSP can outperform all previously proposed dialogue specific and fine-grained pre-training tasks.
Then, we conduct analyses and show that (3) the main source of improvements of DRS come from the training time of TAP in step two, which can even mitigate lack of a good initialization checkpoint, (4) NSP task is crucial to DRS, as opposed to common NLU tasks that can work with only MLM, (5) and the low N-gram train/test overlap \% and low number of distinct N-grams explains why TAP does not improve Douban, and why overfitting occurs for E-commerce.

In short, we make the following contributions: (1) Contrary to previous beliefs, we show that BERT may not be the best initialization checkpoint, and MLM+NSP can outperform all previously proposed dialogue-specific fine-grained learning TAP tasks. (2) We present a set of analyses that identify the source of improvements for TAP, and DRS benchmark characteristics, and (3) contribute a new state-of-the-art on the Ubuntu corpus.

\section{Task-adaptive Pre-training with BERT}

\subsection{Task Formulation}
\label{subsec:task_form}

Given a multi-turn dialogue $c=\{u_1, u_2, ..., u_T\}$ where $u_t$ stands for the $t^{th}$ turn utterance, let $r_i$ denote a response candidate, and $y_i \in \{0, 1\}$ denotes a label with $y_i=1$ indicating that $r_i$ is a proper response for $c_i$. (otherwise, $y_i=0$).  Dialogue Response Selection (DRS) aims to learn a model $f(c, r_i)$ to predict $y_i$.  With the cross-encoder binary classification formulation, DRS shares the exact form as the \textit{Next Sentence Prediction (NSP)} task used in BERT~\cite{devlin-etal-2019-bert}.  

\subsection{Step 1: Initialize from Checkpoint}
\label{subsec:initialization}
All previous works have exclusively used BERT~\cite{devlin-etal-2019-bert} as their initialization checkpoint and does not consider other open-source pre-trained language models.
 We hypothesize this is due to BERT's NSP task and DRS sharing the same task formulation and thus may learn representations that are more helpful to DRS.
However, RoBERTa~\cite{roberta} has shown that BERT is undertrained and removing the NSP task improves downstream task performance, raising questions on whether BERT is the best choice for DRS.

To verify this assumption, we use both BERT and RoBERTa in our experiments.  To the best of our knowledge, we are the first to include RoBERTa for DRS, and that show both achieves similar performance.

\subsection{Step 2: Task-adaptive Pre-training}
\label{subsec:tap}


\subsubsection{Data Augmentation}
\label{subsubsec:data_aug}

All previous works have performed data augmentation to generate more data for \textit{Task-adaptive Pre-training}.  
While several works have devised fine-grained data augmentation methods such as utterance insertion/deletion ~\cite{whang2021ums}, or next session prediction~\cite{Xu2021LearningAE}, we use a standard data augmentation methodology that is commonly used in dialogue literature~\cite{mehri-etal-2019-pretraining,gunasekara-etal-2019-dstc7}.

Given a context-response pair instance $(c=\{u_1, ..., u_{T}\},{r})$ in the original training set, we generate additional $T-1$ context response pairs $\{(c_1, r_1), ..., (c_{T}, r_{T})\}$, where $c_t = \{u_1, ..., u_t\}, r_t = u_{t+1}$, $t \in \{1, ..., T-1\}$ with a total of $T$ pre-training instances.


\subsubsection{Pre-training Task}

While BERT's MLM+NSP objective naturally works as a default TAP choice~\cite{whang20_interspeech}, most recent works have hypothesized that MLM+NSP is not capable of modeling dialogue patterns, and designed dialogue-specific tasks such as incoherence detection~\cite{Xu2021LearningAE}, order shuffling~\cite{whang2021ums}, fine-grained matching~\cite{li2021small}, etc.  
However,~\citet{han-etal-2021-fine} achieved a new cross-encoder state-of-the-art with a simple variant of MLM+NSP, raising questions on whether these dialogue-specific fine-grained tasks actually learn better.

In our experiments, we follow the input representation of~\cite{whang20_interspeech} and use MLM+NSP for TAP and achieve a new state-of-the-art results on Ubuntu.


\subsection{Step 3: Finetuning}
\label{subsec:finetune}

Last, TAP models are fine-tuned on the original datasets on the DRS/NSP task to ensure a fair comparison.
Considering the same task formulation, our MLM+NSP in step 2 can be viewed as multi-task learning with MLM as an auxiliary task and NSP as the primary task~\cite{ruder2017overview}.

\begin{table*}[t]
    \scriptsize 
    \setlength{\tabcolsep}{3pt}
    \centering
    \begin{tabular}{l|c c c|c c c c c c|c c c}
    \toprule 
     \multirow{2}{*}{Model} & \multicolumn{3}{c}{Ubuntu} & \multicolumn{6}{c}{Douban} & \multicolumn{3}{c}{E-commerce} \\
     & $R_{10}@1$ & $R_{10}@2$ & $R_{10}@5$ & MAP & MRR & $P@1$ & $R_{10}@1$ & $R_{10}@2$ & $R_{10}@5$& $R_{10}@1$ & $R_{10}@2$ & $R_{10}@5$ \\
    \midrule
    \multicolumn{13}{c}{Cross-encoders} \\
    \midrule 
    SA-BERT~\cite{SA-BERT} & 0.855 & 0.928 & 0.983 & 0.619 & 0.659 & 0.496 & 0.313 & 0.481 & 0.847 & 0.704 & 0.879 & 0.985 \\
    +HCL~\cite{su-etal-2021-dialogue} & 0.867 & 0.940 & 0.992 & 0.639 & 0.681 & 0.514 & \underline{0.330} & 0.531 & 0.858 & 0.721 & 0.896 & 0.993 \\
    BERT-VFT~\cite{whang20_interspeech} & 0.858 & 0.931 & 0.985 & - & - & - & - & - & - & - & - & -  \\
    UMS$_{BERT+}$~\cite{whang2021ums} & 0.875 & 0.942 & 0.988 & 0.625 & 0.664 & 0.499 & 0.318 & 0.482 & 0.858 & 0.762 & 0.905 & 0.986 \\
    +FGC~\cite{li2021small} & 0.886 & 0.948 & 0.990 & 0.627 & 0.670 & 0.500 & 0.326 & 0.512 & 0.869 & - & - & - \\
    BERT-SL~\cite{Xu2021LearningAE} & 0.884 & 0.946 & 0.990 & - & - & - & - & - & - & 0.776 & 0.919 & 0.991 \\
    BERT-FP~\cite{han-etal-2021-fine} & 0.911 & 0.962 & \underline{0.994} & 0.644 & 0.680 & 0.512 & 0.324 & 0.542 & 0.870 & 0.870 & 0.956 & 0.993 \\
    \midrule 
    \multicolumn{13}{c}{Bi-encoders} \\
    \midrule 
    DR-BERT~\cite{DR-BERT} & 0.910 & 0.962 & 0.993 & \textbf{0.659} & \textbf{0.695} & \textbf{0.520} & \textbf{0.338} & \textbf{0.572} & \textbf{0.880} & \textbf{0.971} & \textbf{0.987} & 0.997 \\
    - w/o CL~\cite{DR-BERT} & 0.888 & 0.943 & 0.988 & 0.616 & 0.655 & 0.487 & 0.309 & 0.501 & 0.819 & 0.891 & 0.955 & 0.991 \\
    \midrule 
    \multicolumn{13}{c}{Ours - TAP} \\
    \midrule 
    BERT + MLM+NSP     & 0.912 & 0.966 & \underline{0.994} & 0.644 & 0.684 & 0.511 & 0.323 & 0.548 & 0.853 & \underline{0.926}	& \underline{0.980} & \textbf{0.998}  \\
    RoBERTa + MLM+NSP  & 0.908 & 0.964 & 0.992 & \underline{0.651} & \underline{0.691} & \textbf{0.520} & 0.329 & \underline{0.563} & 0.867 & 0.916 & 0.976 & \textbf{0.998} \\
    \midrule 
    \multicolumn{13}{c}{Ours - TAP + Fine-tune} \\
    \midrule 
    BERT + MLM+NSP     & \textbf{0.923}    & \textbf{0.969} & \textbf{0.995} & 0.644 & 0.684 & 0.514 & 0.327 & 0.538 & 0.858 & 0.867 & 0.964	& 0.996 \\
    RoBERTa + MLM+NSP  & \underline{0.921} & \underline{0.967} & \underline{0.994} & 0.643 & 0.682 & 0.508 & 0.325 & 0.546 & \underline{0.873} & 0.850 & 0.950 & 0.993 \\
    \bottomrule 
    \end{tabular}
    \caption{Results on response selection benchmarks.  We \textbf{bold-faced} best results and \underline{underline} second-best results.}
    \label{tab:results}
    \vspace{-10pt}
\end{table*}

\section{Experimental Setup}

\subsection{Implementation Details}
\label{subsec:implementation_details}

We used the open-source \textsc{PyTorch-Lightning} framework~\cite{falcon2019pytorch, PyTorch, wolf-etal-2020-transformers} to implement our models.  We use the $BERT_{Base}$ model architecture with the Adam~\cite{Kingma2015AdamAM} optimizer, and performed grid search over learning rates of $\{1e-5, 5e-5, 1e-4\}$ for both TAP and fine-tuning.  For TAP, we trained for 50 epochs and maintained a 1:1 pos:neg sampling ratio for NSP.  For fine-tuning, we average results over 3 random seeds.
More details can be found in the appendix.

\subsection{Evaluation}
\label{subsec:eval}

We follow previous works' protcol and evaluate on three standard response selection benchmarks: (1) Ubuntu corpus~\cite{lowe-etal-2015-ubuntu}, (2) Douban corpus ~\cite{wu-etal-2017-sequential}, and (3) E-commerce corpus ~\cite{zhang-etal-2018-modeling}.  We report recall at 1, 2, and 5 for all 3 sets, and include precision at 1, mean average precision (MAP) and mean reciprocal rank (MRR) for Douban.


\subsection{Baselines}
\label{subsec:baselines}


\paragraph{Cross-encoders}
\textit{Cross-encoders} performs self-attention over the context, response pairs. BERT-VFT~\cite{whang20_interspeech} uses \textit{MLM+NSP} for TAP;  SA-BERT~\cite{SA-BERT} includes speaker embeddings in the input representation; SA-BERT+HCL~\cite{su-etal-2021-dialogue} adopts curriculum learning on top of SA-BERT; UMS$_{BERT}$~\cite{whang2021ums} and BERT-SL~\cite{Xu2021LearningAE} develop dialogue-specific learning tasks, while UMS$_{BERT+}$+FGC~\cite{li2021small} and BERT-FP~\cite{han-etal-2021-fine} uses more fine-grained objectives.  Our model also fall under this category.

\paragraph{Bi-encoders} 
\textit{Bi-encoders} encode context and response separately, resulting in faster inference and allows large number of negative samples per positive instance.  
We report DR-BERT and DR-BERT w/o CL from ~\citet{DR-BERT} that uses MLM for TAP and focuses on step 3.
As DR-BERT uses a pos:neg ratio of 1:63, we include the latter that uses a pos:neg ratio of 1:1 for a fair comparison with our model.

\begin{figure*}[t]
    \centering
    \subfigure[]{\includegraphics[width=0.3\textwidth]{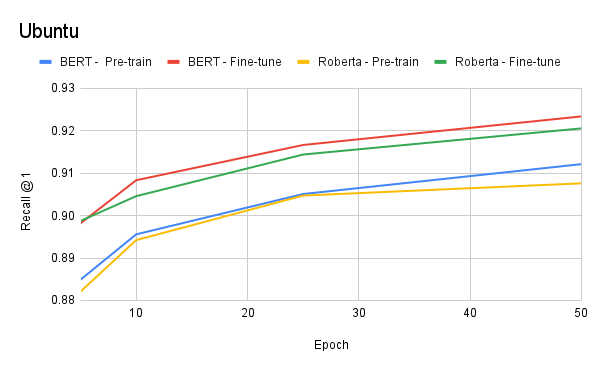}} 
    \subfigure[]{\includegraphics[width=0.3\textwidth]{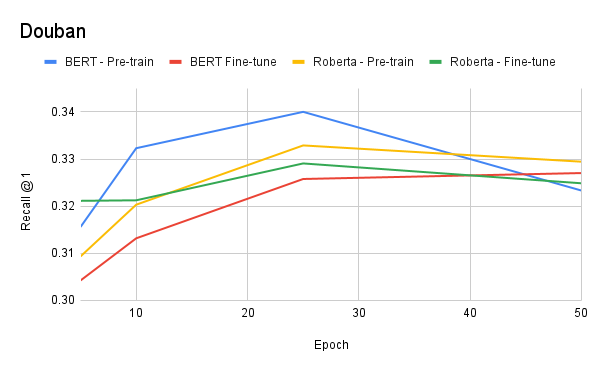}} 
    \subfigure[]{\includegraphics[width=0.3\textwidth]{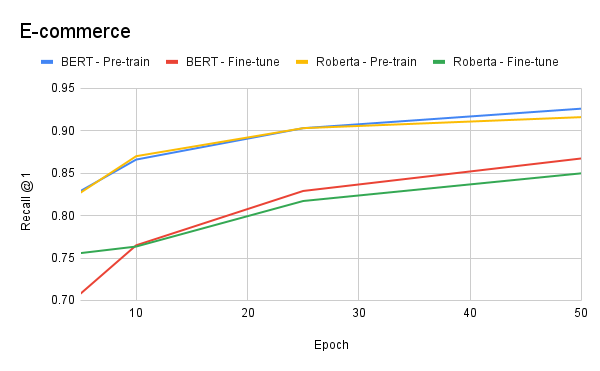}}
    \caption{Test $R_{10}@1$ of (a) Ubuntu, (b) Douban, and (c) E-commerce at 5, 10, 25, and 50 pre-training epochs.}
    \vspace{-10pt}
    \label{fig:pretrainingepochs}
\end{figure*}

\section{Results and Discussion}
\label{sec:results}

\subsection{Main Results}
\label{subsec:main_results}

We report our TAP and fine-tuning results in \Cref{tab:results}.
With MLM+NSP, we achieve a new state-of-the-art (SOTA) result on the Ubuntu corpus, and second-best results on Douban and E-commerce, only falling behind DR-BERT.
This suggests that previously proposed dialogue-specific or fine-grained learning tasks are not necessarily required for these standard DRS benchmarks.  
And to properly evaluate whether these learning tasks capture more fine-grained linguistic patterns, one potential method is to construct adversarial test sets that contain these patterns~\cite{han-etal-2021-evaluation}.

We observe that BERT and RoBERTa resulted in similar performance, with the former outperforming the latter on Ubuntu and E-commerce, while underperforming on Douban. This shows that despite DRS/NSP share the same task formulation, BERT does not always contain more knowledge than RoBERTa for DRS, and other pre-trained models can be explored in the future.

We also notice that fine-tuning achieves better performance for Ubuntu but not for Douban and E-commerce.  We further discuss this in \Cref{subsec:ngram_overlap}.

Though DR-BERT achieves SOTA on Douban and E-commerce, our model still outperforms DR-BERT w/o CL on the latter two, suggesting that the larger negative sampling ratio (NSR) is the key to DR-BERT's superior performance.
Considering that the cross-encoder architecture has to inference each and every context response pair, thus cannot scale NSR efficiently, we leave exploring this hyper-parameter with bi-encoder architectures in future work.

\subsection{Effects of Initialization}
\label{subsec:effects_init}

\begin{figure}
    \centering
    \includegraphics[width=0.4\textwidth]{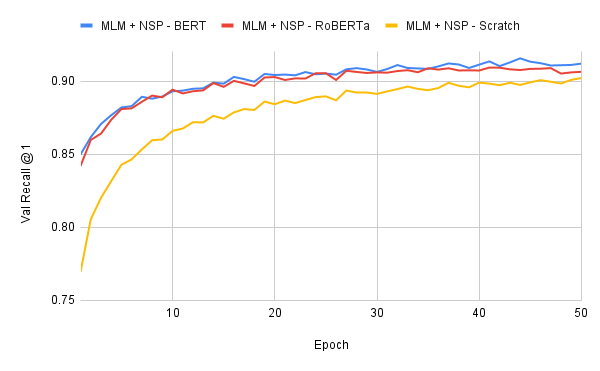}
    \caption{Ubuntu TAP with different initialization.}
    \vspace{-10pt}
    \label{fig:effect_init}
\end{figure}

We analyze the effects of different initialization checkpoint by plotting the Ubuntu validation $R_{10} @ 1$ for (1) BERT, (2) RoBERTa and (3) initialization from scratch in \Cref{fig:effect_init}. 
(1) and (2) perform similarly across training time with the monotonically increasing performance.
While (3) has a large gap with (1) and (2) at the beginning, the gap gradually closes with more training time, indicating that the main source of improvements come from TAP and can mitigate the lack of a pre-trained checkpoint.

\subsection{Effect of TAP Epochs}
\label{subsec:effects_epoch}

We plot both TAP and fine-tuning $R_{10} @ 1$ 
for epochs at 5, 10, 25, and 50 in \Cref{fig:pretrainingepochs}.  
We can see that all 3 datasets exhibit different characteristics.  On the Ubuntu and E-commerce corpus, both TAP and finetuning performance monotonically increases with more epochs.  On the Douban corpus, the performance saturates around 25 epochs for both BERT and RoBERTa, and further training reduces performance.   
This suggests that TAP for longer can improve Ubuntu and E-commerce, but not so for Douban.  We further discuss this in ~\cref{subsec:ngram_overlap}.

\subsection{Effect of TAP Tasks}

\begin{table}[]
    \centering
    \begin{tabular}{c|c|c|c}
    \toprule
     TAP task & MLM+NSP & MLM & NSP \\
    \midrule
      Test $R_{10} @ 1$ & 0.923 & 0.842 & 0.905 \\
    \bottomrule
    \end{tabular}
    \caption{Ubuntu BERT with different TAP tasks.}
    \vspace{-10pt}
    \label{tab:effects_tap_tasks}
\end{table}

We perform TAP with different pre-training tasks and report fine-tuning results on Ubuntu with BERT in \Cref{tab:effects_tap_tasks}.  
We observe that MLM underperforms NSP by a large margin (0.842 vs. 0.905) and multi-tasking MLM+NSP achieves the best results.
This shows that NSP is important for DRS, which is different from common NLU tasks~\cite{wang-etal-2018-glue} where NSP is not helpful~\cite{roberta}.



\subsection{Ngram Analysis}
\label{subsec:ngram_overlap}

We compute overlapping N-grams and discovered that Ubuntu, Douban, and E-commerce's test set had 80.98\%, 9.44\%, and 99.66\% of their 5-grams appearing in their training set, respectively.  Douban's low n-gram \% explains why more TAP does not improve on Douban, and a further inspection shows that Douban's test sets are constructed using corpus from a different domain~\cite{wu-etal-2017-sequential}.  Next, we discovered that despite all having 1 million training instances, Ubuntu, Douban and E-commerce's training set has 25m, 44m, and 900k different 5-grams, respectively.  The lower variation of N-gram in E-commerce is a potential reason why fine-tuning resulted in overfitting.

\section{Conclusion}

We have verified assumptions regarding initialization and pre-training task choices in previous works and achieved a new SOTA on Ubuntu and strong results on Douban and E-commerce.  
Our analyses reveal that the main source of TAP-based DRS comes from more training time, NSP task is crucical for DRS, and that TAP improvements can be estimated by their N-gram overlaps and variations.


\section{Limitations}

From \Cref{subsec:effects_epoch}, we discover that our main source of improvements comes more TAP epochs.  While training for longer monotonically improves performance for Ubuntu and E-commerce, the marginal gains decrease over time, which can be time-consuming and costly.  Another limitation is the cross-encoder formulation that jointly encodes the context-response pairs.  This prevents our model to effectively increase the negative sampling ratio to compare with the bi-encoder architectures such as DR-BERT that has shown superior performance on Douban and E-commerce.  

\bibliography{emnlp2022}
\bibliographystyle{acl_natbib}

\appendix 

\section{Hyperparameter Details}

We use the \textsc{PyTorch-Lightning} framework~\cite{PyTorch, falcon2019pytorch} to implement our models.  

We use pre-trained weights and tokenizers from \textsc{Huggingface} for initialization checkpoints.  For BERT initialization, we use \textsc{bert-base-uncased} for Ubuntu corpus, and \textsc{bert-base-chinese} for Douban and E-commerce.  For RoBERTa initialization, we use \textsc{roberta-base} for Ubuntu, and \textsc{chinese-roberta-wwm-ext} for Douban and E-commerce. 

We trained one run for each TAP model, and use the $BERT_{Base}$ model architecture with $12$ layers, $12$ attention heads, with $768$ as hidden size.

We set batch size to $256$ (except for RoBERTa on Ubuntu that uses $192$), and use the Adam optimizer with learning rate $1e-5$, $\beta_1=0.9$ and $\beta_2=0.99$.
We set gradient normalization to $1.0$ and used $500$ warm up steps with linear schedule.  We set maximum length $L$ of the context response pair with special tokens to 256.  If the length exceeds $L$, we truncate the input by maintaining a 3:1 ratio between context and response as in \citet{han-etal-2021-fine}.  These configurations to both task-adaptive pre-training and fine-tuning to avoid discrepancy between them.

For TAP, we first perform grid search over learning rates of $\{1e-5, 5e-5, 1e-4, 5e-4\}$ and picked the best model based on validation recall and further train it until 50 epochs.   We train only 1 run and use random seed of $1234$.

For fine-tuning, we first perform grid search over the same set of learning rates on BERT/RoBERTa models without TAP, and choose the best learning rate based on recall on the validation set for the TAP models with the same initialization and datasets.  The fine-tuning model keeps all model weights when possible, which includes the NSP head.  We fine-tune for 1 epoch on the original training set and average results over 3 random seeds $1234$, $5678$, and $10111213$.

We used an AWS p3.16x instance and employed mixed-precision~\cite{Micikevicius2018MixedPT} to speed up training.  
Pre-training one epoch on Ubuntu, Douban, and E-commerce corpus took around 1.5 hour, 1 hour, and 1 hour, respectively.

\end{document}